\documentclass[conference]{IEEEtran}
\IEEEoverridecommandlockouts

\usepackage[utf8]{inputenc}  
\usepackage{cite}
\usepackage{amsmath,amssymb,amsfonts}
\usepackage{algorithmic}
\usepackage{graphicx}
\usepackage{textcomp}
\usepackage{xcolor}
\usepackage{url}
\usepackage{booktabs}
\usepackage{comment}
\usepackage{caption}  
\usepackage{subcaption}  
\usepackage{soul}
\usepackage{multirow}
\usepackage{cite}
\usepackage{acronym}
\usepackage{listings}
\makeatletter
\newcommand{\linebreakand}{%
  \end{@IEEEauthorhalign}
  \hfill\mbox{}\par
  \mbox{}\hfill\begin{@IEEEauthorhalign}
}
\makeatother

\begin{document}

\title{Design and testing of an agent chatbot supporting decision making with public transport data\\
}

\author{\IEEEauthorblockN{Luca Fantin}
\IEEEauthorblockA{
\textit{University of Padova}\\
Padova, Italy \\
\textit{Motion Analytica s.r.l.} \\
Mestre, Italy \\
luca.fantin@studenti.unipd.it \\
luca.fantin@motionanalytica.com}
\and
\IEEEauthorblockN{Marco Antonelli}
\IEEEauthorblockA{\textit{beanTech s.r.l.} \\
Udine, Italy \\
marco.antonelli@beantech.it}
\and 
\IEEEauthorblockN{Margherita Cesetti}
\IEEEauthorblockA{\textit{beanTech s.r.l.} \\
Udine, Italy \\
margherita.cesetti@beantech.it}
\linebreakand
\IEEEauthorblockN{Daniele Irto}
\IEEEauthorblockA{\textit{beanTech s.r.l.} \\
Udine, Italy \\
daniele.irto@beantech.it}
\and
\IEEEauthorblockN{Bruno Zamengo}
\IEEEauthorblockA{\textit{Motion Analytica s.r.l.} \\
Mestre, Italy \\
bruno.zamengo@motionanalytica.com}
\and
\IEEEauthorblockN{Francesco Silvestri}
\IEEEauthorblockA{
\textit{University of Padova}\\
Padova, Italy \\
francesco.silvestri@unipd.it}
}

\maketitle

\acrodef{GenAI}[GenAI]{generative artificial intelligence}
\acrodef{LLM}[LLM]{large language model}
\acrodef{GTFS}[GTFS]{General Transit Feed Specification}\
\acrodef{DL}[DL]{Deep Learning}
\acrodef{LOS}[LOS]{level of service}
\acrodef{DBMS}[DBMS]{DataBase Management System}

\begin{abstract}
Assessing the quality of public transportation services requires the analysis of large quantities of data on the scheduled and actual trips and documents listing the quality constraints each service needs to meet. Interrogating such datasets with SQL queries, organizing and visualizing the data can be quite complex for most users. This paper presents a chatbot offering a user-friendly tool to interact with these datasets and support decision making. It is based on an agent architecture, which expands the capabilities of the core Large Language Model (LLM) by allowing it to interact with a series of tools that can execute several tasks, like performing SQL queries, plotting data and creating maps from the coordinates of a trip and its stops.

This paper also tackles one of the main open problems of such Generative AI projects: collecting data to measure the system's performance. Our chatbot has been extensively tested with a workflow that asks several questions and stores the generated query, the retrieved data and the natural language response for each of them. Such questions are drawn from a set of base examples which are then completed with actual data from the database. This procedure yields a dataset for the evaluation of the chatbot's performance, especially the consistency of its answers and the correctness of the generated queries.
\end{abstract}

\section{Introduction}

Public transportation services are an essential part of modern city lives. Therefore, it is fundamental that the agencies responsible for such services are able to intuitively and efficiently analyze the data related to them. Such analyses require the interrogation of SQL databases containing data on the service that are offered and actually provided. However, learning query languages like SQL can be challenging for beginners, and creating elaborate queries can be difficult even for experienced users. Furthermore, directly accessing such databases is often possible only for an authorized pool of users, which is usually smaller than the set of all possible stakeholders of the information contained in the data. Thus, the implementation of a text-to-SQL system \cite{text2sql} can help remove such obstacles and allow a wider audience to perform analysis on such structured data. Another difficulty of interrogating databases comes from the interpretation and visualization of the data, which is as non-trivial and can affect the usefulness of the analysis as much as retrieving the data itself. \Ac{GenAI} can help with this.

This paper proposes a chatbot bringing all these possibilities together. Given a natural language question in input, its core \ac{LLM} analyzes it, transforms it into a SQL query, uses it to retrieve the desired data from a database and returns it in a user-friendly way. This is made possible thanks to the implemented agent architecture. In this approach, the capabilities of the \ac{LLM} are expanded with a series of tools able to perform other operations and interact with the base model; for example, checking the SQL query for syntactic errors, executing it, and using the retrieved data to draw a map of a certain route if the user's question asks so; such tasks are considerably difficult to achieve for the \ac{LLM} on its own.

Along with the description of the application, we also propose a first methodology to quantitively assess its performance, measuring whether the generated query retrieved the right dataset or not. Given a set of question templates, we complete them by inserting specific service data while trying to simulate a random human-user behavior: we change the phrasing, use synonyms and add random errors in the form of incorrect data. We give these questions to the chatbot several times and compare the generated query and retrieved results with the versions we consider correct. We also propose an analysis of this data in both a quantitative and qualitative way.

The paper is structured as follows. In Section \ref{sec:previous} we provide an overview of previous works on AI approaches to public transport data and to text-to-SQL, and on metrics of chatbot's performance. Then, in Section \ref{sec:dataset} we describe the data interrogated by the chatbot, in Section \ref{sec:chatbot} we present the application's architecture and in Section \ref{sec:examples} we show some examples of its use. Finally, in Section \ref{sec:testing} we show how we collect and analyze the data on the system's performance.
\section{Previous work} \label{sec:previous}

The groundwork for this paper was laid down in \cite{mobility_chatbot}. It proposes a similar chatbot operating on mobility data, specifically tourism data, to answer natural language questions on this data. The chatbot displays the retrieved data in text form and produces graphs from it when requested. The different data domain implies a series of differences between such chatbot and the application proposed in this paper.

Furthermore, several papers explored the application of AI in decision making based on public transport data and similar data sources. Paper \cite{iatransport} presents a review of some of these applications. Out of the many considered papers, only a few employ \ac{LLM} or other \ac{DL} techniques, suggesting the application of such methods is still quite novel. Article \cite{chatbottransport} presents a series of AI applications by several public transport agencies. They include chatbots, tailored towards a series of audiences, such as services users and customer service workers.

A number of different text-to-SQL approaches were explored in the literature. Paper \cite{text2sql} presents a review of different systems employing a number of \ac{DL} technologies ranging from LSTM models to \ac{LLM}. Out of these, paper \cite{macsql} is of particular interest. This paper divides the whole text-to-SQL procedure into three sub-tasks handled by different agents: determining the most relevant tables for the question out of the available schema, decomposing the user's question into subquestions and generating SQL queries for each of them, and checking and refining the complete query before returning it to the user. In our application, all these tasks are handled by different tools used by the same agent.

Finally, different solutions were proposed to measure chatbot's performance. For instance, in \cite{answerability}, the authors presents a novel metric and mathematical framework to evaluate conversational AI applications across many different domains. In the proposed example, the chatbot's question-answer pairs were manually labelled depending on the relevance and helpfulness of the answers. In our case, we go beyond the textual answer and evaluate the answers on the basis of the retrieved data used to generate the answers.
\section{Dataset} \label{sec:dataset}

This section presents a generic overview of the database interrogated by the chatbot. It was built from three publicly available datasets containing the information on the public transport services provided by the agencies TPER, for the cities of Bologna and Ferrara and nearby municipalities \cite{gtfs_tper}, and ATM, for the metropolitan city of Milano \cite{gtfs_atm}. Section \ref{sec:base_data} describes these base datasets and Section \ref{sec:processed_data} presents an overview of how they were processed.

\subsection{Base datasets} \label{sec:base_data}

The publicly available datasets comply with the \ac{GTFS} standard \cite{gtfs} for public transport schedules and the associated geographical information. It is made of several text files written as CSVs for the different aspects of the services. They are connected in a similar manner to a relational database, although they cannot be copied as-is in a \ac{DBMS}. Such obstacles, along with the adjustments employed to solve these problems, are discussed in Section \ref{sec:processed_data}.

\begin{figure}[h]
    \centering
    \includegraphics[width=0.8\linewidth]{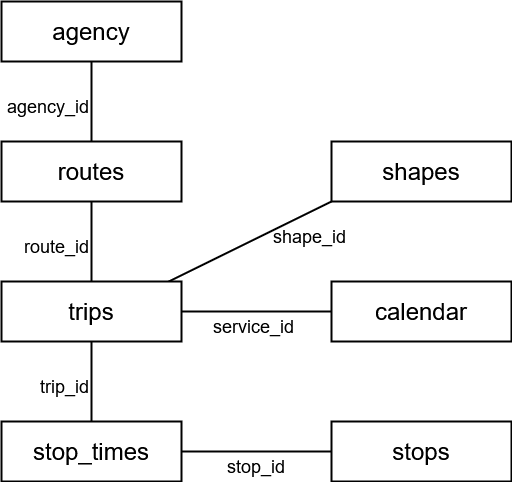}
    \caption{GTFS scheme representing the connections between the most important files (based on an image in Wikimedia Commons); only those mentioned in Section \ref{sec:base_data} are included here. Please note that this represents the conceptual links between the files, and should not be interpreted as an ER diagram.}
    \label{fig:base_gtfs}
\end{figure}

The full list of files allowed by the \ac{GTFS} standard contains 13 different files, most of which are optional. These are the most important files, present in all base datasets used in this application:

\begin{itemize} 
    \item \emph{agency}: generic info about the agency providing the dataset;
    \item \emph{routes}: groups of trips presented as a single service to users; in practice, these are the individual lines offered to service users (for instance, line 11, line 18, etc.)
    \item \emph{trips}: individual travels performed by public transport vehicles (for instance, the line 18 bus departing from the first stop at 8:30 am);
    \item \emph{calendar}: service dates for routes specified by a weekly schedule with start and end dates; for example, the set of all trips performed by line 18 during weekdays between September 2024 and June 2025 are associated to a single service;
    \item \emph{shapes}: sequence of geographical points (longitude - latitude pairs) representing the path of a trip on a map;
    \item \emph{stops}: locations where service users can be picked up and dropped off;
    \item \emph{stop\_times}: times where trips arrive at and depart from stops.
\end{itemize}

As shown in Fig. \ref{fig:base_gtfs}, these files reference each other to fully describe the public transport service, using fields such as \emph{agency\_id}, \emph{route\_id} or \emph{trip\_id} in a similar manner to foreign keys in a proper relational database.

\subsection{Processing} \label{sec:processed_data}

The datasets were processed in order to remove the obstacles preventing them to be uploaded to the database. For example, each record of the \emph{shapes} file is uniquely identified by the id of the specific shape and the position of a certain point in the full sequence, but \emph{trips} only contains the shape id. Thus, the \emph{shapes} file was decomposed into three tables: one for the shapes, one for the individual points and one storing the sequences of points, referencing the other two tables.

Furthermore, a single GTFS dataset describes the service of a single agency; to store GTFS data from multiple agencies in a single database, the original text files had to be expanded. The id of the agency was added to the primary key of every table, to correctly identify all records.

Finally, other data sources were added to the database; for instance, the list of all Italian municipalities with their respective geometries, published by the Italian institute of statistics (ISTAT) and publicly available online \cite{istat_municipalities}. This allows to associate each entry in the \emph{stops} table to the municipality containing a specific stop. Such procedure can be repeated, enabling future developments and data enrichments.

\section{Chatbot} \label{sec:chatbot}

This section presents a high-level technical description of the chatbot. Section \ref{sec:architecture} describes the general architecture, while \ref{sec:implementation} provides more details on its implementation and the used libraries.

\subsection{Architecture} \label{sec:architecture}

The chatbot implements an \emph{agent} architecture \cite{agents_google}. Although there is not a widely accepted formal definition of this concept yet, it can be described as an application with the ability to carry out a task by `reasoning' on it, preparing a step-by-step plan to solve it and executing said plan exploiting a series of diverse skills. Its central component is a \ac{LLM}, serving as the centralized decision maker. The agent can use a series of \emph{tools} able to perform tasks that would otherwise be challenging or even infeasible for the \ac{LLM} on its own. All of this is brought together by an \emph{orchestration layer}, which contains the reasoning capabilities, memory and general scope of the agent. Such methodology greatly expands the basic capabilities of language models, opening a wide array of possible applications.

In our case, the main ability needed by the chatbot is to interrogate a database. Given the user's question, the agent passes it to the LLM, which generates a SQL query following a set of rules and example queries specified in the model's prompt. Once the query is generated, this is provided as input to a series of tools that can execute it or verify its correctness. The \ac{LLM} then uses the retrieved records to generate a natural language response to the user's initial question. If the user requires a map of a certain route, the agent can then invoke another tool that can process the data retrieved by the database and build the map.

\subsection{Implementation} \label{sec:implementation}

\begin{figure}[h]
    \centering
    \includegraphics[width=\linewidth]{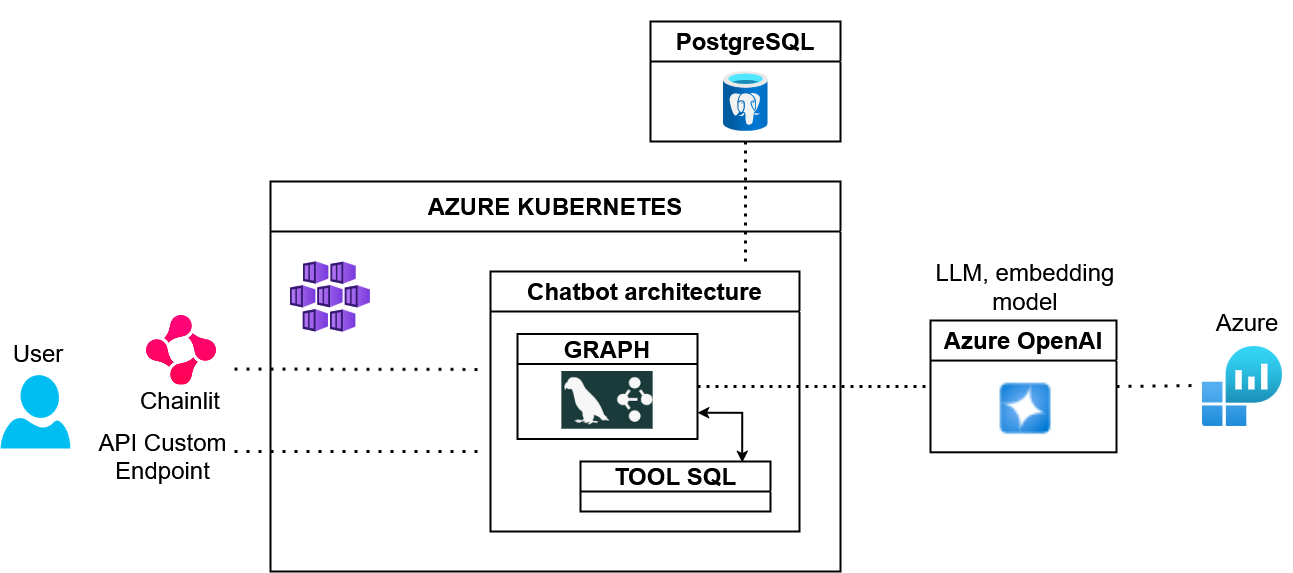}
    \caption{High-level representation of the architecture implemented by the chatbot.}
    \label{fig:scheme}
\end{figure}

Our chatbot uses a number of Python libraries to handle different aspects. A high-level diagram representing the implementation can be seen in Fig. \ref{fig:scheme}.

The agent architecture is implemented using the \emph{Langchain} library \cite{langchain}. It is designed to streamline the development of conversational AI applications in all its aspects, like integrating various services and resources, orchestrating the process and generating responses. In particular, it was used to create the tools that communicate with a PostgreSQL database, using the \emph{SQLAlchemy} library.

The resources used by the application are provided by the cloud services of Microsoft Azure and its environment \emph{Azure AI}, which gives access to the GPT-4-Turbo \ac{LLM}, currently considered state-of-the-art, and the \emph{text-embedding-ada-002} model \cite{embedding} employed to build the embeddings used to find the most relevant example queries for any input question.

The front-end is handled by the \emph{Chainlit} library. It is designed to bring conversational AI applications to a production-ready status. Many popular libraries and frameworks, like the aforementioned Langchain, can be integrated and made available for use through a user-friendly UI.

Finally, the application is deployed using Docker and Kubernetes. This gives access to the chatbot both through a web browser using the Chainlit interface and by calling a custom web API.

\section{User experience} \label{sec:examples}

\lstset{language=XML,basicstyle=\ttfamily\footnotesize,breaklines=true,showspaces=false,showstringspaces=false}
\begin{figure*}[h]
  \centering
  \begin{lstlisting}
    <prompt>
      <task>You are an AI assistant designed to...</task>
      <database>
        <table>
          <name>agency</name>
          <description>agencies responsible for the public transportation services reported in this database</description>
          <definition>CREATE TABLE agency...</definition>
          <comments>
            COMMENT ON column agency.agency_id IS 'unique identifier of the agency';
            COMMENT ON column agency.agency_name IS 'full name of the agency';
            ...
          </comments>
        </table>
        ...
        <foreign_keys>
          ALTER TABLE routes ADD CONSTRAINT route_managed_by FOREIGN KEY (agency_id) REFERENCES agency(agency_id);
          ...
        </foreign_keys>
      </database>
      <rules>
        <rule>Query only relevant columns.</rule>
        <rule>If a query returns nothing, report the empty result.</rule>
        <rule>Always double check your query.</rule>
        ...
      </rules>
    </prompt>
    \end{lstlisting}
  \caption{Simplified version of the textual prompt given to the agent's core LLM.}
  \label{fig:prompt}
\end{figure*}
\lstset{basicstyle=\ttfamily\small,breaklines=true,showspaces=false,showstringspaces=false}

This section presents a few examples of interactions between a user and the chatbot, showcasing some examples of questions that can be asked and the reasoning and planning skills of our application. We
start with a simple case, asking the number of routes provided by the agency of
Bologna:

\begin{lstlisting}
  Q: How many routes are managed by the agency of Bologna?
  A: The agency of Bologna manages 238 routes.
\end{lstlisting}

The user's question is first given as input to the \ac{LLM} to generate the SQL
query. The model is guided with a textual prompt, written with an XML-like
structure that uses ad-hoc tags \cite{prompts}. The prompt is divided into
several sections, providing a general description of the task, a complete
description of the database and a set of rules to follow. The database is fully
specified, with DDL statements to create the tables, description for every
table and column, and the statements creating the foreign key relationships. A simplified example of the prompt can be seen in Fig. \ref{fig:prompt}.


This prompt engineering approach yields several advantages compared to a
plain text approach, like better separating and organizing the various
sections. This reflects the results of more large-scale studies comparing
several prompt engineering approaches on different applications \cite{prompts}.
Furthermore, the examples that are considered most relevant to the user's
questions are attached at the end of the prompt.

The \ac{LLM} can now generate the query to find the number of routes managed by the agency of Bologna:
\begin{lstlisting}
select count(distinct r.route_id) from routes r join agency a using (agency_id) where upper(a.agency_hq_city) like upper('%Bologna%')
\end{lstlisting}
Next, the agent invokes the different tools as needed. First, the generated
query is passed to a tool that looks for syntactical and logical errors and
corrects them when necessary. This is achieved by using the \ac{LLM} with a
different prompt, containing common errors. Then, another tool is used to
execute the query and retrieve the needed data from the database. Thus, we get the answer:
\begin{lstlisting}
[(238,)]
\end{lstlisting}
Finally, the language model generates the final answer using the data returned
by the query: 
\begin{lstlisting}
The agency of Bologna manages 238 routes.
\end{lstlisting}

The chatbot can also handle questions with multiple conditions:
\begin{lstlisting}
  Q:  Which routes managed by the agency of Ferrara perform the most trips between  7:00 and 9:00 on working days?
  A:  The routes managed by the agency of Ferrara that perform the most trips between 7:00 and 9:00 on working days are the following:
      <list of 10 routes with name, ends of line and average number of trips>
      These are just some examples of routes with the highest number of trips in that hour range.
\end{lstlisting}

The generated SQL query includes several logical conditions, regarding service
validity days and hour ranges for instance, and mathematical aggregate
operations, to count the average number of trips for every involved route.
Furthermore, this example showcases some of the interpretation rules of the
dataset. For example, we did not specify how to aggregate the data: whether to consider the trips done in a specific date range, on a specific working day or just consider all working days recorded in the database; and whether to compute the sum of all considered trips, or the average number of trips among the considered time period. In such cases, the chatbot has to make these choices by itself, following the instructions included in the prompt. In this first version, the data only includes services that are active on the current date.
One future improvement can consider every working day and compute the average
number of trips over all working days, if the question does not specify a date
range. The chatbot will then have to specify all such assumptions made to
answer the question, so that the user can correctly interpret the answer.

Finally, the chatbot can correctly retrieve the data needed to build route
maps. For this, the SQL tool interrogates a view, a virtual table including all
coordinates of stops and shape points for all routes. Such data is then passed
to a tool that handles its processing and the creation of an interactive map
that shows the route shape and its stops.

\begin{figure}[h]
  \centering
  \includegraphics[width=0.9\linewidth]{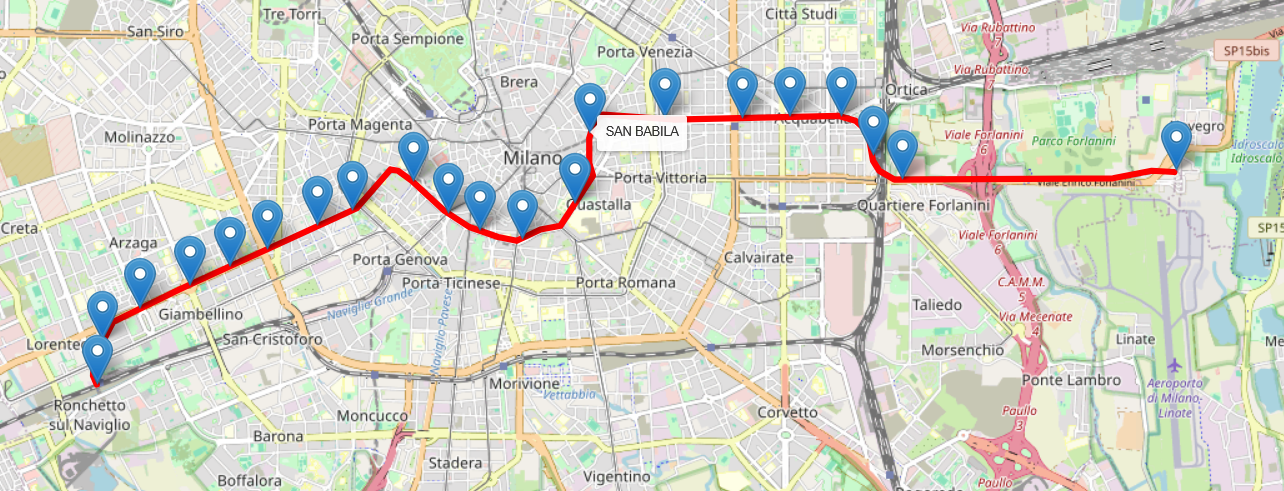}
  \caption{Example of the maps built by the dedicated tool.}
  \label{fig:map}
\end{figure}

Fig. \ref{fig:map} shows a prototype of such maps, representing one of the
metropolitan lines of Milan. The shape is displayed on top of a Leaflet map.
The markers represent the stops, and their names are displayed upon hovering on
them with the cursor. However, work still needs to be done to correctly display
them in Chainlit.

\section{Performance analysis} \label{sec:testing}

This section presents our methodology to assess the performance of the chatbot by repeatedly asking a series of questions, storing the generated queries and retrieved records for each of them and comparing this data with the reference data. Section \ref{sec:methodology} presents this procedure in greater detail and sections \ref{sec:results1} and \ref{sec:results2} report and analyze the results.

\subsection{Methodology} \label{sec:methodology}

The test questions were prepared on the basis of three templates:

\begin{enumerate}
    \item \emph{Which routes serve a certain municipality?}
    \item \emph{Which municipalities are served by a certain route?} 
    \item \emph{What is the average number of trips that belong to a certain route and use a certain stop?} 
\end{enumerate}

A version of ChatGPT based on the GPT-4o \ac{LLM} was used to automatically complete these questions while exploiting the linguistic knowledge encoded in the training set of the model. A total of 42 questions was generated by inserting actual data from the Bologna dataset. Moreover, additional conditions were appended to a subset of these questions, to test how well the core \ac{LLM} is able to recognize them and modify the example queries appended to the model's prompt accordingly. These conditions namely require an answer related to specific hour ranges, date ranges, sets of week days or directions (inbound / outbound). Finally, purposefully incorrect data was occasionally used to determine if the chatbot can recognize it. For instance, certain questions generated by template 3 ask for data regarding a route that does not actually use the specified stop.

\begin{figure}[h]
    \centering
    \includegraphics[width=\linewidth]{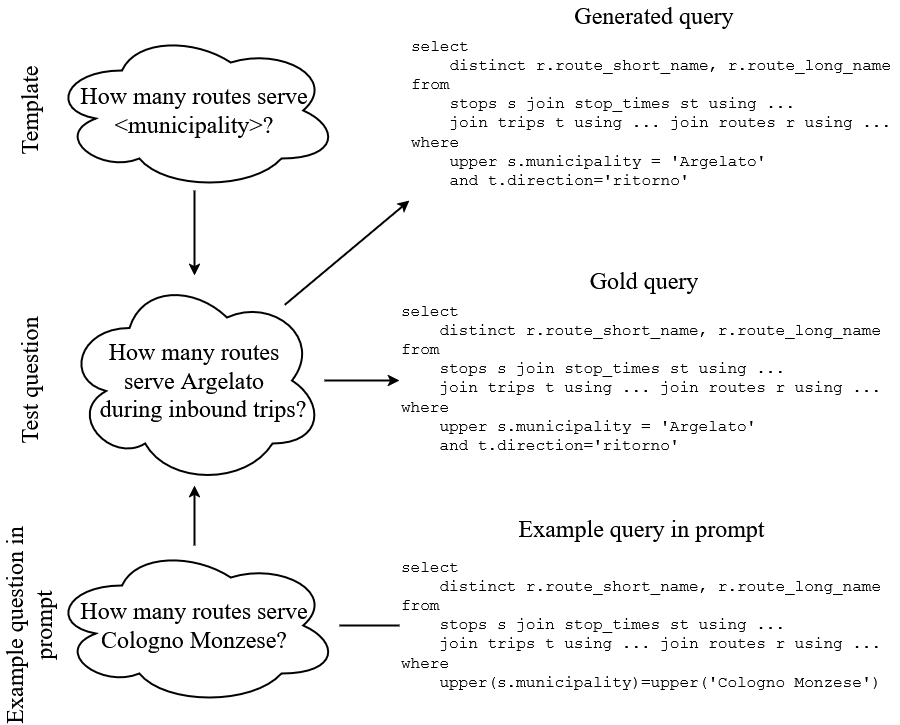}
    \caption{Example of a test question, with its original template, corresponding example question in the SQL tool's prompt and example, gold and generated queries.}
    \label{fig:test_question}
\end{figure}

This question set was then given as input to a Python script that repeatedly asked these questions to the chatbot via its API and stored the results for each of them in a dedicated database, for a total of 146 questions made to the chatbot. Additionally, a \emph{gold query} was manually written and executed for each question in the test set to obtain reference data to compare the chatbot's responses to. An example comprising all these components is shown in Fig. \ref{fig:test_question}.

Questions 1 and 2 are easier to answer, requiring a simple list of entries (municipalities and routes, respectively) satisfying a set of conditions. Thus, for them, the list returned by the generated and golden queries are compared to determine the number of common and different elements between them. On the other hand, question 3 is more complex, demanding further computations on the retrieved records (number of trips) that return a single value in the end. This suggests a comparison based on the computation of the difference between the values computed by the chatbot and by the gold query.

\subsection{Templates 1-2 results} \label{sec:results1}

Of all the questions asked to the chatbot, 112 were generated from templates 1 and 2. For 27 of these, the chatbot generated an incorrect query. These errors can be divided into two types:
\begin{itemize}
    \item 10 cases: SQL syntax errors, which lead the chatbot to not return any results and answer to the user that it does not know the answer to the question;
    \item 17 cases: incorrect results returned by syntactically correct queries, returning lists of stops or routes instead of a list of municipalities.
\end{itemize}
The most common syntactic errors involve the incorrect use of table aliases and the comparison of data of different types. For example, the table \emph{trips} has a column indicating its direction, whose only valid values are the strings \emph{andata} (outbound) and \emph{ritorno} (inbound). However, in several cases the generated query assumed that this column has integer data, with 0 for outbound and 1 for inbound; this is how the direction is specified in the \ac{GTFS} standard.

For the remaining 85 questions, the result sets returned by the generated and gold queries were compared with a Python script, to compute the number of records present in both sets or only one of them. This was performed by turning the result sets into Pandas DataFrames, ordering the columns alphabetically to ensure the correctness of the analysis and comparing the data using the \texttt{merge} method. The results of this analysis can be summarized as follows:
\begin{itemize}
    \item 59 cases: the generated and gold queries return the exact same result set;
    \item 2 cases: the generated query returned all entries retrieved by the gold query, together with other incorrect results (false positive rates: 83\% and 75\%);
    \item 10 cases: the generated query returns only a portion of the result set retrieved by the gold query (average percentage of gold records not retrieved by the generated query, i.e. false negative rate: 40\%);
    \item 14 cases: the result sets returned by the two queries are disjoint.
\end{itemize}

Overall, the chatbot provided the correct answers for around 53\% of the questions.

\subsection{Template 3 results} \label{sec:results2}

The remaining 34 questions asked to the chatbot were generated from template 3. The correct results were returned in 6 cases, all of which asked the same two questions. In these instances, the agent picked the correct example query and modified it accordingly. The remaining 28 cases fall into two types similar to those presented in Section \ref{sec:results1}. Most of the syntactically correct queries belonging to the second category return values equal to 0, with which the \ac{LLM} produces an answer saying that there are no trips. This happens for both questions with valid and invalid route-stop pairs. While such an answer is correct from the user's point of view for the latter type of questions, it was not generated using a correct query.


\section{Conclusion}

In this paper we presented a chatbot supporting decision-making processes based on public transport data with a novel approach. It is based on the agent architecture, which allows a core \ac{LLM} to execute a task by reasoning and executing a plan which invokes a series of tools performing other sub-tasks. Along with a few qualitative examples to showcase the user experience, we also proposed a methodology to assess the performance of our application in an automated and numerical way. The results show that while the chatbot is good at recognizing relevant examples and modifying them accordingly to the user's questions, it still struggles to build queries that require bigger modifications and adaptations of such examples. This analysis opens up the road for several possible uses and improvements of the chatbot.

In our current line of work, we are repurposing the agent architecture following a multi-agent approach. The several sub-tasks needed by the chatbot are each executed by a single agent, and all resulting components are connected as a graph, with edges encoding how these agents interact and exchange data with each other. This allows to separate and optimize tasks more easily, as well as improving the overall control flow. As a result, we expect this migration to allow a considerable improvement of the accuracy of the chatbot's answers and the creation and visualization of plots and maps. This new architecture is being implemented using the \emph{Langgraph} Python library. 

Another open issue concerns the integration of new data sources, such as real-time transport data and service documents. For instance, real-time data allows final users of public transport service to obtain information on the lines they need to use, and comparing GTFS and real-time data allows the computation of the \ac{LOS} of the public transport services, a measure of the quality of the services. 
\section*{Acknowledgements}

The authors would like to thank Alessandro Nalin from University of Bologna for his support in interpreting and processing the \ac{GTFS} data, and in getting a better understanding of the public transport services.

\bibliographystyle{IEEEtran}
\bibliography{references}

\begin{thebibliography}{10}
\providecommand{\url}[1]{#1}
\csname url@samestyle\endcsname
\providecommand{\newblock}{\relax}
\providecommand{\bibinfo}[2]{#2}
\providecommand{\BIBentrySTDinterwordspacing}{\spaceskip=0pt\relax}
\providecommand{\BIBentryALTinterwordstretchfactor}{4}
\providecommand{\BIBentryALTinterwordspacing}{\spaceskip=\fontdimen2\font plus
\BIBentryALTinterwordstretchfactor\fontdimen3\font minus \fontdimen4\font\relax}
\providecommand{\BIBforeignlanguage}[2]{{%
\expandafter\ifx\csname l@#1\endcsname\relax
\typeout{** WARNING: IEEEtran.bst: No hyphenation pattern has been}%
\typeout{** loaded for the language `#1'. Using the pattern for}%
\typeout{** the default language instead.}%
\else
\language=\csname l@#1\endcsname
\fi
#2}}
\providecommand{\BIBdecl}{\relax}
\BIBdecl

\bibitem{text2sql}
X.~Zhu, Q.~Li, L.~Cui, and Y.~Liu, ``Large language model enhanced text-to-sql generation: A survey,'' \url{https://arxiv.org/abs/2410.06011}, 2024.

\bibitem{mobility_chatbot}
L.~Padoan, M.~Cesetti, L.~Brunello, M.~Antonelli, B.~Zamengo, and F.~Silvestri, ``Mobility chatbot: supporting decision making in mobility data with chatbots,'' in \emph{2024 25th IEEE International Conference on Mobile Data Management (MDM)}, 2024, pp. 295--300.

\bibitem{iatransport}
A.~Jevinger, C.~Zhao, J.~Persson, and P.~Davidsson, ``Artificial intelligence for improving public transport: a mapping study,'' \emph{Public Transport}, vol.~16, pp. 1--60, 11 2023.

\bibitem{chatbottransport}
{UITP Union Internationale des Transports Publics}, ``Ai in public transport,'' \url{https://www.uitp.org/publications/ai-public-transport/}, Mar. 2025.

\bibitem{macsql}
B.~Wang, C.~Ren, J.~Yang, X.~Liang, J.~Bai, L.~Chai, Z.~Yan, Q.-W. Zhang, D.~Yin, X.~Sun, and Z.~Li, ``Mac-sql: A multi-agent collaborative framework for text-to-sql,'' \url{https://arxiv.org/abs/2312.11242}, 2025.

\bibitem{answerability}
P.~Gupta, A.~A. Rajasekar, A.~Patel, M.~Kulkarni, A.~Sunell, K.~Kim, K.~Ganapathy, and A.~Trivedi, ``Answerability: A custom metric for evaluating chatbot performance,'' in \emph{Proceedings of the 2nd Workshop on Natural Language Generation, Evaluation, and Metrics (GEM)}.\hskip 1em plus 0.5em minus 0.4em\relax Abu Dhabi, United Arab Emirates (Hybrid): Association for Computational Linguistics, Dec. 2022, pp. 316--325.

\bibitem{gtfs_tper}
Tper, ``Open data,'' \url{https://solweb.tper.it/web/tools/open-data/open-data.aspx}.

\bibitem{gtfs_atm}
ATM, ``Pubblicazione orari del trasporto pubblico locale in formato gtfs,'' \url{https://www.amat-mi.it/it/servizi/pubblicazione-orari-trasporto-pubblico-locale-formato-gtfs/}.

\bibitem{gtfs}
{Mobility Data}, ``General transit feed specification,'' \url{https://gtfs.org/}.

\bibitem{istat_municipalities}
ISTAT, ``Confini delle unità amministrative a fini statistici,'' \url{https://www.istat.it/notizia/confini-delle-unita-amministrative-a-fini-statistici-al-1-gennaio-2018-2/}.

\bibitem{agents_google}
J.~Wiesinger, P.~Marlow, and V.~Vuskovic, ``Agents,'' \url{https://www.kaggle.com/whitepaper-agents}, February 2025.

\bibitem{langchain}
``Langchain,'' \url{https://python.langchain.com/docs/introduction/}.

\bibitem{embedding}
OpenAI, ``New and improved embedding model,'' \url{https://openai.com/index/new-and-improved-embedding-model/}, Dec. 2022.

\bibitem{prompts}
L.~Pawlik, ``How the choice of llm and prompt engineering affects chatbot effectiveness,'' \emph{Electronics}, vol.~14, p. 888, 02 2025.

\end{thebibliography}

\end{document}